\documentclass[10pt]{article}

\usepackage[accepted]{eiml_icml2026} 
\usepackage{amsmath,amssymb}
\usepackage{booktabs}
\usepackage{graphicx}
\usepackage{hyperref}
\usepackage{multirow}
\usepackage{enumitem}

\icmltitlerunning{LLM Doesn't Know What It Doesn't Know}

\begin{document}

\twocolumn[
\icmltitle{LLM Doesn't Know What It Doesn't Know: Detecting Epistemic Blind Spots via Cross-Model Attribution Divergence on Clinical Tabular Data}


\begin{icmlauthorlist}
\icmlauthor{Akshat Dasula}{inst1,inst2}
\icmlauthor{Prasanna Desikan}{inst1}
\icmlauthor{Jaideep Srivastava}{inst2}
\end{icmlauthorlist}

\icmlaffiliation{inst1}{Centific AI Research}
\icmlaffiliation{inst2}{Department of Computer Science \& Engineering, University of Minnesota-Twin Cities, Minneapolis, USA}

\icmlcorrespondingauthor{Akshat Dasula}{dasul001@umn.edu}

\vskip 0.1in

\begin{abstract}

Large language models (LLMs) are increasingly applied to structured clinical data, yet whether they can recognize the limits of their own knowledge on such tasks remains unexplored. We study this question through the lens of cross-model attribution divergence with the goal of reducing epistemic uncertainty for structured tasks, comparing Qwen~2.5~7B and XGBoost on a prediction task via attribution divergence analysis. We report four findings. First, LLM verbalized confidence is \textit{epistemically vacuous}, it outputs a near-constant ($0.856$--$0.937$) regardless of whether accuracy is $49\%$ or $75.3\%$, tracking prompt format rather than prediction quality. Second, the LLM exhibits an \textit{inverse difficulty effect}: accuracy drops to $64.8\%$ when XGBoost is $99\%$ correct, but matches XGBoost ($73.8\%$ vs.\ $73.1\%$) when it is moderately uncertain. Third, few-shot examples and SHAP-derived feature evidence are \textit{orthogonal, super-additive} interventions: they reduce the Attribution Disagreement Score (ADS) from $1.54$ to $0.38$ and improve accuracy from $49\%$ to $75.3\%$ without training. Fourth, a cross-model calibrator that determined LLM reliability using attribution divergence signals reduces expected calibration error from $0.254$ to $0.080$, replacing uninformative verbalized confidence with patient-specific reliability estimates, without accessing model internals or requiring repeated inference. We frame these findings as a cold start problem for LLMs on structured data and outline a path toward genuine epistemic self-awareness.

\end{abstract}
]

\printAffiliationsAndNotice{}

\section{Introduction}
Large language models (LLMs) have demonstrated remarkable capabilities across medical question answering, clinical note summarization, and diagnostic reasoning \cite{singhal2023, nazi2024}. This success has driven growing interest in applying LLMs to structured clinical prediction tasks, using electronic health record (EHR) data to predict patient outcomes such as disease onset, deterioration, and readmission. Yet a persistent empirical finding complicates this trajectory, tree-based models such as XGBoost consistently outperform LLMs on tabular prediction tasks \cite{brown2025large, grinsztajn2022tree, shwartz2022tabular}, often by substantial margins. Prior work has documented this performance gap across clinical datasets and LLM families \cite{brown2025large}, but has largely treated it as a single aggregate number, the LLM achieves a lower AUROC, without examining the epistemic structure beneath it.
This gap raises a question that is fundamental to the safe deployment of LLMs in clinical workflows: \textit{does the LLM know when it doesn't know?} If an LLM produces a confident but incorrect prediction for a critically ill patient, and offers no signal that its reasoning is unreliable, the consequences are qualitatively different from those of a model that flags its own uncertainty. Reliable uncertainty estimation is not merely desirable, it is a prerequisite for clinical deployment, regulatory compliance, and responsible integration into human decision-making.

We show that on structured clinical tabular data, the LLM's epistemic self-awareness is entirely absent and that this absence has identifiable, addressable structure. Studying acute kidney injury (AKI) prediction on MIMIC-IV \citep{johnson2023mimiciv}, we compare the reasoning of XGBoost (AUROC =0.85) with Qwen 2.5 7B Instruct \cite{qwen25}  across four systematically varied prompting conditions. Rather than treating the performance gap as a monolithic failure, we decompose it into three epistemic dimensions: \textit{what} the LLM attends to (measured via attribution divergence between the LLM's self-reported feature importance and XGBoost's SHAP-derived attributions), \textit{whether} it knows when it is wrong (measured via analysis of verbalized confidence), and \textit{where} its failures concentrate (measured via uncertainty stratification by structured model confidence). We do not argue that LLMs should replace structured models for clinical tabular prediction. Rather, we address the practical reality that LLMs are increasingly deployed alongside structured models in clinical workflows and often without rigorous evaluation of their reasoning, reliability, or epistemic limitations. Our findings are as follows:
\begin{itemize}
\item \textbf{Confidence invariance.} We show that LLM verbalized confidence \citep{lin2022teaching, tian2023just} is determined by prompt template, not prediction quality, producing a near-constant for zero-shot and few-shot conditions, regardless of performance. This extends prior findings of LLM overconfidence \citep{xiong2023can} to a stronger claim, on clinical tabular data, verbalized confidence carries zero information about correctness.

\item \textbf{Inverse difficulty effect.} The LLM fails most when XGBoost is most confident but matches XGBoost in its uncertain zone. This reveals complementary failure modes: the LLM struggles on data-specific distributional patterns inaccessible through pretraining, but adds genuine value where general medical knowledge is sufficient.

\item \textbf{Super-additive attribution alignment.} Few-shot examples and SHAP-derived feature evidence \citep{shap} are orthogonal interventions that address independent dimensions of the LLM's reasoning gap. Their combined effect on attribution alignment exceeds the sum of their individual effects, improving the performance without any parameter updates.

\item \textbf{Cross-model calibration.} We construct a lightweight calibrator using cross-model attribution divergence features that reduces expected calibration error producing patient-specific reliability estimates without accessing LLM internals or requiring repeated inference.
\end{itemize}

We frame these findings as characterizing a \textit{cold start problem} for LLMs on structured data, the LLM possesses relevant medical knowledge but lacks direction (which features to attend to) and self-assessment (how confident to be). The structured model already deployed in the clinical workflow can provide both. Our results establish the diagnostic foundation for a longer-term research:  using cross-model calibration targets as reward signals for reinforcement learning, ultimately teaching LLMs to internalize genuine epistemic self-awareness across tasks.

\vspace{-10pt}
\section{Related Work}
\textbf{LLMs on clinical tabular data.}~~The application of LLMs to structured EHR data has received growing attention as hospitals explore integrating language models into existing clinical workflows. \citet{brown2025large} conducted a systematic comparison of GPT-3.5 and GPT-4 against gradient-boosted trees on clinical prediction tasks using MIMIC-IV and Vanderbilt University Medical Center data, finding that LLMs are substantially less effective across performance, calibration, and fairness metrics. \citet{hegselmann2023tabllm} explored few-shot prompting strategies for tabular classification, demonstrating that serialization format and example selection significantly affect LLM performance on structured data. \citet{yildiz2025will} reviewed the broader landscape of LLMs for clinical prediction, identifying poor calibration, limited external validation, and high infrastructure costs as persistent barriers. These works focus on comparing \textit{prediction performance} between LLMs and structured models. Our work complements this line of research by comparing \textit{reasoning processes}, examining not just whether the LLM gets the answer right, but whether it attends to the same features and recognizes when its reasoning is unreliable.

\textbf{LLM uncertainty estimation.}~~Reliable uncertainty quantification is essential for deploying LLMs in safety-critical settings, yet remains an open challenge. \citet{xiong2023can} proposed a systematic framework for black-box confidence elicitation, benchmarking prompting strategies, sampling methods, and aggregation techniques across five LLMs and five datasets. Their key finding is that LLMs tend to be overconfident, potentially imitating human patterns of expressing confidence, established verbalized confidence as a problematic but widely-used uncertainty signal. \citet{heo2024llms} further demonstrated that verbalized confidence is influenced by task formatting rather than actual correctness, raising questions about its suitability for downstream decision-making. \citet{lin2022teaching} introduced the concept of teaching models to express uncertainty in words, and \citet{tian2023just} proposed elicitation strategies for RLHF-tuned models, finding that prompting techniques can partially mitigate overconfidence but do not eliminate it. We extend these findings to clinical tabular prediction and demonstrate a qualitatively stronger failure mode, on structured EHR data, LLM verbalized confidence is not merely miscalibrated but entirely \textit{invariant} to prediction quality, a constant determined by prompt template that carries zero information about correctness.

\textbf{Attribution disagreement.}~~The disagreement problem in explainable AI is well-documented, different feature attribution methods applied to the same model frequently produce conflicting importance rankings \cite{krishna2022disagreement}. This body of work has primarily focused on comparing methods (SHAP vs LIME) within a single model. More recently, cross-model attribution comparison has emerged as a research direction. Work in financial tabular classification found that LLM and LightGBM feature attributions exhibit directional agreement \cite{finai2025}, establishing that LLMs and tree-based models reason about fundamentally different features even when predicting the same outcome. We extend cross-model attribution comparison to clinical data and, go beyond documenting the divergence: we use the resulting disagreement signals as features for uncertainty estimation and calibration. To our knowledge, this is the first work to repurpose attribution divergence as a model-external uncertainty signal.

\textbf{Calibration and selective prediction.}~~Expected calibration error (ECE) \cite{guo2017calibration} is the standard metric for assessing whether a model's stated confidence matches its actual accuracy. Traditional post-hoc calibration methods such as temperature scaling and Platt scaling require access to model logits, which may be unavailable for closed-source LLMs or impractical in deployment. Conformal prediction \cite{vovk2005conformal} provides distribution-free coverage guarantees as a post-hoc wrapper but does not modify the model's internal confidence or self-assessment, the model remains unaware of its own unreliability. Selective prediction approaches allow models to abstain on uncertain inputs, improving accuracy on the accepted subset at the cost of coverage. Our approach is complementary to all of these: we construct a model-external calibrator that uses cross-model attribution signals to produce calibrated reliability estimates, requiring neither access to LLM internals nor repeated inference. This positions it as a lightweight alternative suitable for real-time clinical deployment where computational cost and model access constraints are practical considerations.

\vspace{-10pt}
\section{Methods}

\subsection{Data and Task}
Following the methodology of \cite{aki}, we replicate the Acute Kidney Injury (AKI) prediction task on the MIMIC-IV clinical database \citep{johnson2023mimiciv}, where labels are derived from a rolling-window application of the KDIGO criteria to serum creatinine measurements. After filtering encounters according to the original study's criteria to reach a population of $\sim$209,000 hospital admissions, we extracted a balanced preliminary cohort of 10,000 encounters (5,000 per class) to evaluate binary classification (AKI: Stage 1,2,3 vs. No AKI). Each admission is represented by 321 features, including categorized vital signs, demographics, Clinical Classifications Software (CCS) diagnosis codes, and discretized laboratory values, notably, labs are encoded as low, normal, high, or unknown to capture clinical ordering patterns alongside physiological values. For our experimental pipeline, we split 80/20 into train ($n=8{,}000$) and test ($n=2{,}000$) to build our XGBoost model, with a 15\% LLM evaluation subset ($n=300$) drawn from the test set for preliminary results.



\subsection{Structured Model: XGBoost}
An XGBoost classifier  was built on the train-test split, achieving an AUROC~$= 0.88$, accuracy~$= 0.83$. To accommodate LLM context constraints and ensure input simplicity, we reduced the feature space to the top 50 most influential variables, the model's performance remained unchanged after re-fitting this subset, confirming that predictive integrity was maintained. On the 300-sample subset designated for LLM evaluation, the model achieved an AUROC $\approx 0.85$ and an accuracy $\approx 0.823$. For each encounter in this subset, we recorded the model's predicted label, confidence score, and the top five local feature attributions, including their importance scores and directional impact extracted via SHAP \citep{shap}. This model will be used to provide grounding and as a reference for the subsequent LLM experiments.

\subsection{LLM Experiments}
To address the privacy concerns inherent in sensitive healthcare data, we conducted our preliminary experiments using a locally hosted Qwen 2.5 7B Instruct model \cite{qwen25}. We evaluated the model on the 300-sample test set across four experimental conditions: (1) \textbf{Zero-shot (ZS)}, utilizing serialized patient features with a prediction prompt; (2) \textbf{ZS + SHAP}, where the prompt was augmented with the XGBoost model’s top-5 SHAP features, their importance scores and directions; (3) \textbf{Few-shot (FS)}, incorporating four class-balanced labeled examples; and (4) \textbf{FS + SHAP}, which combined few-shot learning with SHAP-based feature injection. In all settings, the model was required to generate a binary prediction of AKI outcome, a verbalized confidence \cite{lin2022teaching, tian2023just} score (ranging from $0$ to $1$), and a ranking of the top five features with their associated risk directions. To ensure consistent and parseable results, we enforced structured JSON output using Jsonformer \citep{jsonformer}.

\subsection{Attribution Disagreement Score (ADS)}
To quantify the alignment between models, we define an Attribution Disagreement Score (ADS). Let $R_A = (f_1, \dots, f_K)$ and $R_B = (g_1, \dots, g_K)$ represent the top-$K$ feature rankings from XGBoost's SHAP and the LLM, respectively. To account for disjoint sets, any feature present in the union $\mathcal{U} = R_A \cup R_B$ but absent from a specific model's top-$K$ list is assigned a tie-breaking rank of $K+1$. The ADS is calculated as:
\begin{equation}
\text{ADS} = 1 - \tau(R_A, R_B)
\end{equation}
where $\tau$ denotes Kendall’s rank correlation coefficient \cite{kendall1938}, yielding a score from $0$ (perfect agreement) to $2$ (perfect anti-correlation). We further supplement this with Jaccard overlap \cite{jaccard1912}, top-1 match rates, and directional agreement (sign-consistency) on shared features. While rank correlation is a baseline for comparing XAI explainers \citep{krishna2022disagreement}, our application of this metric to estimate uncertainty via \textit{cross-model-family} attribution comparison is, to our knowledge, a novel approach for validating LLM reasoning.

\subsection{Cross-Model Calibrator}

We train calibrators via cross-validation to predict P(\text{is LLM correct}) from: ADS, Jaccard overlap, directional agreement, top-1 match, XGBoost confidence, LLM confidence, and their absolute difference. The calibrator's output replaces the LLM's verbalized confidence as a reliability estimate. We evaluate using expected calibration error (ECE) and AUROC for discriminating correct from incorrect LLM predictions

\section{Results}
\subsection{Performance Progression}
Table~\ref{tab:performance} summarizes performance, attribution divergence, and calibration across all conditions. The zero-shot LLM performs at chance (accuracy=0.49, F1=0.000), Few-shot prompting restores meaningful prediction (accuracy=0.683, F1=0.596), demonstrating that the LLM possesses implicit clinical reasoning capacity that requires task demonstrations to activate. The combination of few-shot examples with SHAP injection achieves 75.3\% accuracy and F1=0.722, closing over half the gap to XGBoost (82.3\%) without any parameter updates. The improvement from zero-shot to few-shot + SHAP is 0.263 (95\% CI: [0.200,0.327], bootstrap), confirming statistical significance.

\begin{table}[t]
\caption{Performance, attribution divergence, and calibration across conditions ($n=300$). S denotes SHAP injection.}
\label{tab:performance}
\vskip 0.1in
\centering
\small
\begin{tabular}{@{}lccccc@{}}
\toprule
 & \textbf{XGB} & \textbf{ZS} & \textbf{ZS+S} & \textbf{FS} & \textbf{FS+S} \\
\midrule
Accuracy & .823 & .490 & .520 & .683 & \textbf{.753} \\
F1 & .795 & .000 & .133 & .596 & \textbf{.722} \\
AUROC & .851 & .500 & .529 & .688 & \textbf{.756} \\
\midrule
ADS $\downarrow$ & --- & 1.536 & 0.881 & 1.395 & \textbf{0.378} \\
Jaccard $\uparrow$ & --- & .145 & .658 & .204 & \textbf{.741} \\
Top-1 $\uparrow$ & --- & .023 & .377 & .010 & \textbf{.820} \\
Dir.\ Agr.\ $\uparrow$ & --- & .600 & .999 & .441 & \textbf{1.00} \\
\midrule
Confidence & --- & .856 & .856 & .937 & .939 \\
ECE $\downarrow$ & --- & .366 & .336 & .254 & \textbf{.202} \\
\bottomrule
\end{tabular}
\vskip -0.1in
\end{table}

\subsection{Finding 1: LLM Confidence Is Epistemically Vacuous}
LLM verbalized confidence \citep{lin2022teaching, tian2023just} is determined entirely by prompt format, not prediction quality. Zero-shot conditions produce a confidence of 0.856 regardless of whether SHAP evidence is injected (accuracy 49\% vs 52\%). Few-shot conditions produce 0.937 regardless of SHAP injection (accuracy 68.3\% vs 75.3\%). Confidence does not vary between correct and incorrect predictions, nor between easy and hard patients. On the zero-shot condition, confidence predicts errors at exactly chance level (accuracy=0.490, AUROC=0.50). This finding extends prior observations of LLM overconfidence \cite{xiong2023can} to a stronger claim, on clinical tabular data, verbalized confidence is not merely miscalibrated but \textit{completely invariant} to prediction quality, rendering it unsuitable as an uncertainty signal for any downstream decision-making.

\subsection{Finding 2: Inverse Difficulty Effect}
Table~\ref{tab:stratification} reveals a counterintuitive relationship between data difficulty and LLM performance. When XGBoost is highly confident (\textgreater0.85) and nearly always correct (99.0\% accuracy), the few-shot LLM achieves only 64.8\% accuracy. Conversely, when XGBoost is moderately uncertain (confidence 0.70--0.85), the LLM matches it at 73.8\% vs 73.1\%. This \textit{inverse difficulty effect} indicates that the two models have complementary failure modes. XGBoost excels on cases dominated by data-specific distributional patterns learned during training. The LLM performs competitively where these patterns are weak and general medical knowledge is sufficient for prediction. Critically, LLM confidence remains constant across all strata (0.934), confirming the LLM has no mechanism to distinguish cases it can handle from those it cannot.

This finding has a positive implication, the LLM is not uniformly inferior. It adds genuine value in XGBoost's uncertain zone, suggesting that a combined system could leverage the strengths of both models.

\begin{table}[t]
\caption{Few-shot LLM performance stratified by XGBoost confidence level.}
\label{tab:stratification}
\vskip 0.1in
\centering
\small
\begin{tabular}{@{}lcccc@{}}
\toprule
\textbf{XGB Conf.} & \textbf{$n$} & \textbf{XGB Acc.} & \textbf{LLM Acc.} & \textbf{LLM Conf.} \\
\midrule
Low (0.5--0.7) & 35 & .743 & .543 & .934 \\
Med.\ (0.7--0.85) & 160 & .731 & \textbf{.738} & .935 \\
High (0.85--1.0) & 105 & .990 & .648 & .943 \\
\bottomrule
\end{tabular}
\vskip -0.1in
\end{table}

\subsection{Finding 3: Attribution Alignment Is Super-Additive}
The attribution divergence analysis reveals that the LLM's reasoning gap has two orthogonal, independently addressable dimensions. ADS decreases across conditions with a super-additive combined effect:
\begin{itemize}[nosep,leftmargin=1.5em]
\item ZS $\to$ FS: $\Delta$ADS $= 0.141$ (task comprehension)
\item ZS $\to$ ZS+SHAP: $\Delta$ADS $= 0.655$ (feature guidance)
\item ZS $\to$ FS+SHAP: $\Delta$ADS $= 1.158$ (both)
\end{itemize}
The combined reduction (1.158) substantially exceeds the sum of individual reductions (0.141+0.655=0.796), confirming that task comprehension and feature guidance address distinct epistemic gaps that interact constructively when combined.

The nature of each intervention is revealing. Few-shot examples barely change \textit{which} features the LLM attends to (Jaccard: 0.145→0.204) but dramatically improve \textit{how} it maps features to predictions (F1: 0.000→0.596). SHAP injection redirects feature attention effectively (Jaccard: 0.145→0.658, direction agreement: 0.600→0.999) but the LLM exhibits what we term \textit{shallow adoption}: it copies the structured model's attributions verbally while barely changing its predictions (accuracy improves only 3 points). The LLM agrees that certain features increase risk but does not adjust its conclusion accordingly. Only when both interventions are combined does the LLM achieve both aligned reasoning (ADS =0.378, Jaccard =0.741, top-1 match =82.0\%, direction agreement =100\%) and strong performance (75.3\%).

This decomposition has practical significance, it demonstrates that for the current clinical deployment, both task demonstrations and structured model evidence should be provided to LLMs. Neither alone is sufficient, and their combination yields benefits beyond what either provides independently.

\subsection{Finding 4: Cross-Model Signals Enable Calibration}


The attribution divergence signals that reveal the LLM's reasoning failures also power an effective calibration system.Table~\ref{tab:calibration} compares calibrators trained on cross-model features on the few-shot condition. A logistic regression achieves the lowest ECE (0.043) but modest discrimination (AUROC =0.551). An XGBoost calibrator with Platt scaling achieves substantially stronger discrimination (AUROC=0.71) with ECE=0.080, a 68.5\% reduction from the raw LLM's ECE of 0.254. This confirms that nonlinear interactions between attribution divergence features carry predictive signal showing that with more information calibrator can distinguish reliable predictions from unreliable ones.

The raw LLM outputs 0.937 for all patients while achieving 68.3\% accuracy, producing a 25.4-point overconfidence gap. The calibrated XGBoost replaces this uninformative constant with patient-specific reliability scores that achieve 77\% accuracy in predicting whether the LLM will be correct on a given patient. This result demonstrates that the structured model serves a dual role: it provides both a diagnostic perspective (identifying where the LLM's reasoning diverges) and a calibration signal (predicting when those divergences will cause errors), without accessing LLM internals or requiring repeated inference.


\begin{table}[t]
\caption{Cross-model calibration on the few-shot condition.}
\label{tab:calibration}
\vskip 0.1in
\centering
\small
\begin{tabular}{@{}lcc@{}}
\toprule
\textbf{Calibrator} & \textbf{AUROC} & \textbf{ECE} \\
\midrule
Raw LLM confidence & 0.533 & 0.254 \\
Logistic Regression & 0.551 & 0.043 \\
XGBoost + Platt scaling & \textbf{0.710} & \textbf{0.080} \\
\bottomrule
\end{tabular}
\vskip -0.1in
\end{table}

\subsection{Complementarity}
An oracle analysis confirms complementary failure modes across models. Of 300 patients, both models are correct on 65.3\%, XGBoost alone on 17.0\%, the few-shot LLM alone on 3.0\% (9 patients), and neither on 14.7\%. The oracle ensemble achieves 85.3\% versus XGBoost's 82.3\%, a 3-point gain attributable entirely to the 9 patients where the LLM captures clinical signal that XGBoost misses. These patients have notably low XGBoost confidence (mean=0.649), confirming that the LLM's unique contributions concentrate in the structured model's uncertain zone.

A simple confidence-threshold router does not capture this complementarity. Selective prediction does not improve over XGBoost alone. This indicates that while the complementary signal exists, more sophisticated routing mechanisms are needed to exploit it. Developing such mechanisms, potentially using attribution divergence features or learned routing policies, represents a promising direction for future work.

\section{Discussion}

\subsection{The Cold Start Problem}
Our results reveal that applying an LLM to a structured prediction task resembles a cold start problem. The LLM possesses substantial medical knowledge, as evidenced by its competitive performance in XGBoost's uncertain zone (73.8\% vs 73.1\%) but lacks two critical capabilities when confronted with raw tabular data, \textit{direction} (which features to attend to) and \textit{self-assessment} (how confident to be in its prediction). Without task demonstrations, the LLM cannot map serialized features to predictions (F1=0.000). Without structured model evidence, it attends to the wrong features (Jaccard =0.145). Without either, its confidence is a meaningless constant (0.856) that reflects prompt formatting rather than epistemic state. The LLM's knowledge is real but inaccessible, locked behind an inability to orient itself within the feature space and an inability to recognize when its orientation is wrong.

The progressive alignment results demonstrate that this cold start is addressable. Few-shot examples provide direction by teaching the feature-to-prediction mapping. SHAP injection provides orientation by pointing the LLM toward relevant features. Their super-additive interaction ($\Delta$ADS =1.158\textgreater 0.796) indicates these are independent dimensions of the cold start, solving one does not solve the other, but solving both yields compounding benefits. The cross-model calibrator addresses the self-assessment gap externally, reducing ECE from 0.254 to 0.080 while achieving AUROC=0.710 by replacing the LLM's uninformative confidence with a score that reflects actual reliability. We propose that this cold start framing generalizes beyond clinical prediction. Any setting where an LLM is applied to structured, domain-specific tabular data will likely exhibit the same pattern, latent reasoning capacity, misaligned feature attention, and absent confidence calibration. The structured models already deployed in these domains, can serve the same dual role we demonstrate here, providing both the direction (via attribution injection) and the calibration signal (via cross-model divergence features) that the LLM lacks.

\subsection{Epistemic Implications}
The confidence invariance finding carries implications for the broader study of epistemic intelligence in AI systems. Existing work on LLM uncertainty \cite{xiong2023can, tian2023just} has documented overconfidence as a \textit{magnitude} problem, the LLM reports 90\% when it should report 70\%. Our finding is qualitatively different, the LLM's confidence carries \textit{zero information} about correctness (accuracy=49\% on zero-shot). This is not miscalibration in the conventional sense. It is a complete absence of epistemic self-awareness, the LLM cannot distinguish what it knows from what it does not know.

The inverse difficulty effect further sharpens this picture. The LLM's epistemic blind spot is not random, it is precisely identifiable. The LLM fails most on cases where the structured model is most confident, because those cases are driven by distributional patterns learned from training data that are fundamentally inaccessible through pretraining on medical text. The structured model's confidence directly measures whether a prediction requires such distributional knowledge, making it an external epistemic signal.

\subsection{Limitations}

This is a work-in-progress study with several important limitations. We evaluate a single LLM (Qwen 2.5 7B) on a single clinical prediction task (AKI) with a modest evaluation sample (n=300). Our prompting strategy are limited we do not investigate other strategies like chain-of-thought prompting. LLM feature attributions are self-reported rather than computed, introducing potential faithfulness concerns. We are actively pursuing clinician validations for this and for the calibrator as it is dependent on a model's confidence. Injecting SHAP-derived features risks indirect label leakage, though the LLM's top-1 agreement of only 82\% and accuracy gap with XGBoost (75.3\% vs 82.3\%) suggest it does not simply copy the structured model's output, further analysis is needed to ground this. We view these limitations as appropriate for a preliminary study establishing the diagnostic framework and key findings. Future work will address each through multi-model evaluation, multi-task extension, larger cohorts, and clinician evaluation.

\section{Future Work}
This preliminary work establishes the diagnostic foundation for a broader research aimed at teaching LLMs to produce calibrated, task-aware confidence estimates across structured prediction tasks. We outline several directions, spanning immediate applications of the current framework to longer-term research goals.

\textbf{RL-based confidence internalization.}~~The cross-model calibrator produces reliability scores P(\text{LLM correct}) that can serve as training targets for reinforcement learning. By defining a reward signal that penalizes the gap between the LLM's verbalized confidence and the calibrator's output, an RL training loop (e.g., DPO) could teach the LLM to internalize calibrated uncertainty. After training, the LLM would output high confidence when its reasoning aligns with data-driven evidence and low confidence otherwise, eliminating the need for the external calibrator at inference time, letting the LLM to communicate its epistemic state.

\textbf{Multi-task reliability estimation.}~~A natural extension is training a stronger cross-model calibrator across multiple tasks. If a single calibrator predicts LLM reliability across tasks, it would demonstrate that the LLM's epistemic limitations are systematic rather than task-specific. The critical generalization test follows: after RL-based confidence training on multiple tasks, does the LLM's confidence remain calibrated on an \textit{unseen} task with no RL training? Success would indicate genuine epistemic self-awareness; failure would reveal that calibration is task-specific and requires per-task supervision.


\textbf{Formal uncertainty quantification.}~~We plan to compare ADS-based reliability estimation against established uncertainty quantification methods such as, semantic entropy \citep{kuhn2023semantic}, conformal prediction \citep{vovk2005conformal} and token logprob entropy \cite{zhang2025tokur, ma2025estimating} across multiple tasks. If it achieves comparable calibration at a fraction of the computational cost, it would establish cross-model attribution divergence as a practical, lightweight alternative for settings where model internals are unavailable or repeated inference is prohibitively expensive. Combining calibrated internal confidence with conformal external guarantees could yield an architecture for safety deployments.

\bibliographystyle{plainnat}
\bibliography{references}

\end{document}